\documentclass[12pt]{article}
\usepackage[margin=1in]{geometry}
\usepackage{graphicx}
\usepackage{booktabs}
\usepackage{multirow}
\usepackage{amsmath}
\usepackage{url}
\usepackage{hyperref}

\title{RadOnc-GPT: An Autonomous LLM Agent for Real-Time Patient Outcomes Labeling at Scale}

\author{
Jason Holmes$^{1}$, Yuexing Hao$^{1}$, Mariana Borras-Osorio$^{2}$, \\ 
Federico Mastroleo$^{2}$, Santiago Romero Brufau$^{2}$, Valentina Carducci$^{2}$, \\ 
Kathryn M Van Abel$^{2}$, David M Routman$^{2}$, Andrew Y. K. Foong$^{2}$, \\ 
Liv M Muller$^{2}$, Satomi Shiraishi$^{2}$, Daniel K Ebner$^{2}$, \\ 
Daniel J Ma$^{2}$, Sameer R Keole$^{1}$, Samir H Patel$^{1}$, \\ 
Mirek Fatyga$^{1}$, Martin Bues$^{1}$, Brad J Stish$^{2}$, Yolanda I Garces$^{2}$, \\ 
Michelle A Neben Wittich$^{2}$, Robert L Foote$^{2}$, Sujay A Vora$^{1}$, \\ 
Nadia N Laack$^{2}$, Mark R Waddle$^{2}$, Wei Liu$^{1}$\thanks{Corresponding author. Email: \texttt{liu.wei@mayo.edu}} \\
\\
$^{1}$Mayo Clinic, Phoenix, AZ, USA \\
$^{2}$Mayo Clinic, Rochester, MN, USA
}

\date{\today}

\begin{document}
\maketitle

\begin{abstract}
\noindent\textbf{Purpose:} Manual labeling limits the scale, accuracy, and timeliness of patient outcomes research in radiation oncology. We present \textit{RadOnc-GPT}, an autonomous large language model (LLM)–based agent capable of independently retrieving patient-specific information, iteratively assessing evidence, and returning structured outcomes. Our evaluation explicitly validates RadOnc-GPT across two clearly defined tiers of increasing complexity: (1) a structured quality assurance (QA) tier, assessing the accurate retrieval of demographic and radiotherapy treatment plan details, followed by (2) a complex clinical outcomes labeling tier involving determination of mandibular osteoradionecrosis (ORN) in head-and-neck cancer (HNC) patients and detection of cancer recurrence in independent prostate and HNC cohorts requiring combined interpretation of structured and unstructured patient data. The QA tier establishes foundational trust in structured-data retrieval, a critical prerequisite for successful complex clinical outcome labeling.

\noindent\textbf{Methods:} A wrapper program called \textit{LLM Task Streaming} sequentially supplied patient IDs and specific task instructions to RadOnc-GPT. Initially, RadOnc-GPT completed the structured QA task using demographic and radiotherapy treatment plan course data exclusively. In the second tier, RadOnc-GPT autonomously combined structured electrical health record (EHR) data with unstructured clinical notes, radiology, and pathology reports to determine clinical outcomes. Ground-truth labels for ORN and cancer recurrence were retrospectively generated by expert radiation oncologists. Discrepancies between RadOnc-GPT's outputs and ground-truth labels underwent adjudication by independent radiation oncologists, classifying each discrepancy into one of three categories: model error, ground-truth error, or indeterminate.

\noindent\textbf{Results:} Across 895 unique patients, RadOnc-GPT matched all six demographic fields for all 500 patients in the QA tier (100\%) and accurately reproduced radiation-course counts in 497/500 (99.4\%), demonstrating robust retrieval fidelity. In the complex patient outcomes labeling tier, initial accuracy improved significantly post-adjudication: for ORN determination (233 patients) accuracy rose from 84.5\% to 95.2\%, prostate cancer recurrence detection (80 patients) improved from 92.5\% to 95.0\%, and HNC recurrence detection (82 patients) improved from 92.7\% to 96.3\%. Among 48 initial discrepancies, adjudication revealed 30 (63\%) previously unrecognized ground-truth errors, 13 genuine model errors, and 5 indeterminate cases.

\noindent\textbf{Conclusions:} RadOnc-GPT reliably retrieves foundational structured data and subsequently generalizes complex clinical outcome labeling tasks, notably using a single cancer recurrence detection prompt across multiple disease sites. Its high recall performance minimizes clinically critical false negatives, while its dual capacity as labeler and auditor significantly enhances registry data integrity. Thus, RadOnc-GPT enables scalable, trustworthy, and real-time curation of radiation-oncology research datasets.
\end{abstract}

\section{Introduction}

Large-language models (LLMs) have progressed rapidly from general-purpose chatbots to highly specialized systems capable of matching or surpassing human experts across multiple clinical tasks\cite{RN1787, RN1790, RN1169, RN1789, RN1559, RN1553, RN1389, RN1788, RN1445, RN1793}. Specifically, in radiation oncology GPT-4 has attained expert-level performance on radiation-oncology physics exams\,\cite{holmes2023evaluating,wang2024recent}, broad clinical knowledge assessments\,\cite{ramadan2025evaluating,longwell2024performance,dennstadt2024exploring}, clinical task management such as in-basket response\,\cite{hao2025retrospective}, automated drafting of payer-appeal documents\,\cite{kiser2024large}, and harmonizing naming conventions of anatomical structures\,\cite{holmes2024benchmarking}. In addition, recently open source LLMs have been fine-tuned to improve the performance of important clinical tasks in radiation oncology \cite{RN1791}. Further improvements have been demonstrated through retrieval-augmented generation (RAG) approaches, which significantly enhance accuracy in clinical trial screening applications\,\cite{doi:10.1056/AIoa2400181,PMID:37576126,RYBINSKI2024104734,ferber2024endtoendclinicaltrialmatching,gupta2024prismpatientrecordsinterpretation}.

Parallel studies in oncology-specific natural language processing (NLP) have demonstrated the successful extraction of detailed clinical information such as pathology findings\,\cite{Geevarghesejcp-2024-209669}, computed tomography (CT) imaging phenotypes\,\cite{doi:10.1148/radiol.231362}, key entities from discharge letters\,\cite{SIEPMANN2025100378}, toxicity assessments\,\cite{khanmohammadi2024novel}, patient comorbidities\,\cite{info:doi/10.2196/58457}, outcomes from lung-ablation procedures\,\cite{GEEVARGHESE2024}, and patient-reported outcomes\,\cite{doi:10.1200/CCI.23.00258}. Recent multi-modal models that integrate imaging and textual data have further expanded capabilities to tasks like tumor contouring\,\cite{oh2024llm} and survival prediction\,\cite{cancers16132402}. However, a significant limitation of most existing NLP approaches and registry curation efforts is their inherent assumption that institutional registry labels represent reliable gold standards. Recent audits, such as a multi-year review of an academic HNC registry, suggest otherwise, revealing significant latent errors, with up to half of true recurrence events potentially missing from automated flags\,\cite{SUTTON2024225}. Correcting these errors via traditional manual review requires substantial staffing resources, highlighting the need for more efficient, scalable methods for auditing and curating clinical outcomes data.

To address these critical limitations, we developed \textbf{RadOnc-GPT}, a GPT-4o–based autonomous agent capable of retrieving structured and unstructured clinical data directly from institutional databases, iteratively assessing the evidence, and synthesizing structured clinical outcomes. In order to evaluate RadOnc-GPT, we designed a two-tier strategy differing by level of complexity: first, a relatively simple structured data retrieval task aimed at establishing the agent's structured information retrieval capability. Because this task involves reproducing structured data, manual review is not required and therefore makes this evaluation step a good candidate for quality assurance purposes. After demonstrating reliable structured-data reproduction, we increase the complexity of the tasks, evaluating RadOnc-GPT's ability to mix structured and unstructured patient data towards performing the task. These high-complexity tasks included determination of mandibular osteoradionecrosis (ORN) in HNC patients and cancer recurrence detection in separate prostate and HNC patient cohorts. A noteworthy design feature of our evaluation was using an identical cancer recurrence detection method (identical input prompt) across two disease sites, prostate cancer and HNC, testing RadOnc-GPT’s ability to generalize clinical reasoning across disease sites.

Through this staged evaluation approach, we aim to answer three key questions:

\begin{enumerate}
    \item Can an autonomous LLM-based agent reliably retrieve and interpret structured clinical data from institutional databases?
    \item Once baseline structured retrieval accuracy is confirmed, can the agent successfully perform complex clinical outcomes labeling using both structured and unstructured data?
    \item Can such an agent simultaneously identify latent errors in existing institutional registry labels, thus serving a dual function of patient outcome labeling and real-time data auditing?
\end{enumerate}

Answering these questions clearly and rigorously establishes autonomous LLM agents as a potentially transformative solution for near-real-time clinical outcomes labeling at scale.

\section{Methods}

\subsection{RadOnc-GPT System Design}
\begin{figure}[ht]
  \centering
  \includegraphics[width=.99\textwidth]{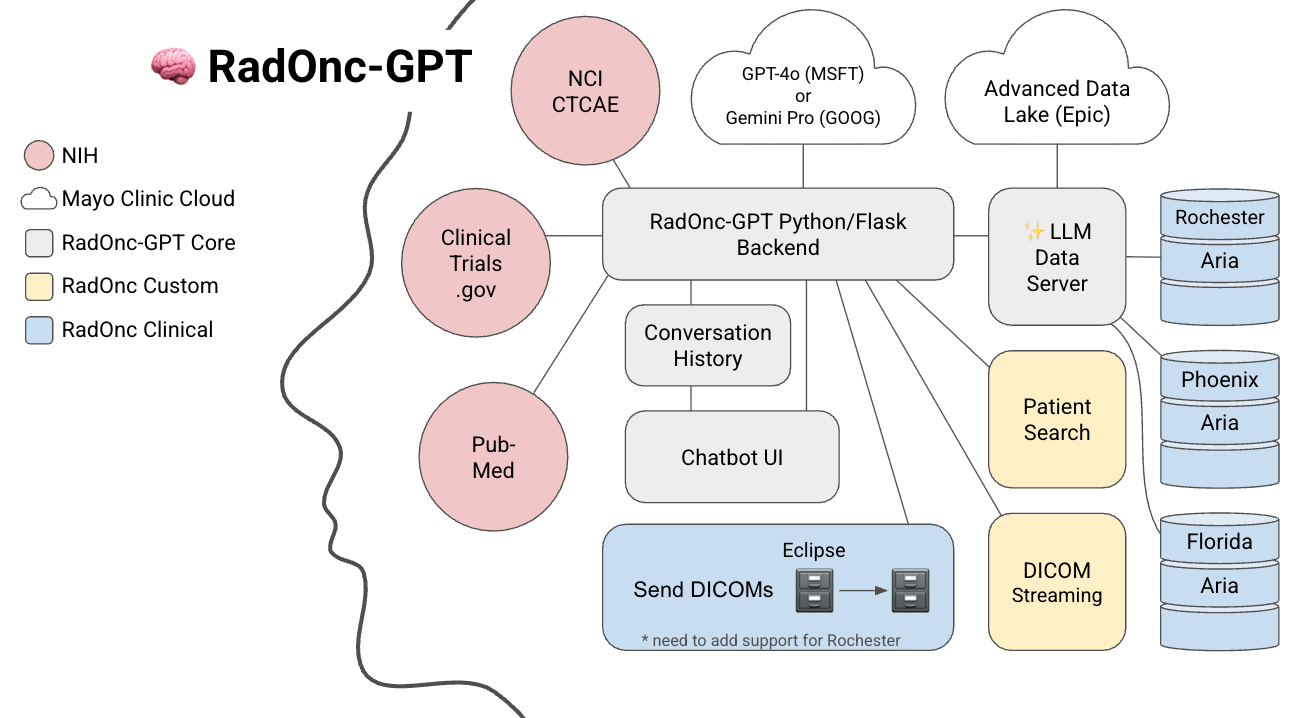}
  \caption{\textbf{RadOnc-GPT Agent Architecture.} }
  \label{fig:radonc_arch}
\end{figure}

Figure~\ref{fig:radonc_arch} illustrates the architecture of RadOnc-GPT, which integrates both internal and external data resources. Internal connections include Mayo Clinic’s radiation oncology–specific database system (Minnesota, Arizona, Florida), Aria (Varian Medical Systems, Palo Alto, CA), and enterprise electronic health record (EHR) database systems interfacing with Epic (Epic Systems, Verona, WI). RadOnc-GPT further connects with public data resources including PubMed, ClinicalTrials.gov, and the National Cancer Institute (NCI) Common Terminology Criteria for Adverse Events (CTCAE), each connected via their public APIs. RadOnc-GPT is designed as an LLM-based agent, defined here as a system capable of conducting multi-turn conversations while autonomously determining which functions to call and when to stop responding. The RadOnc-GPT system makes use of the many data sources by calling upon whitelisted functions that retrieve, process, and return data to GPT-4o (OpenAI). The specifications of these functions—including their inputs and outputs—are explicitly detailed within the system prompt so as to promote good decision making on the part of the LLM in deciding which functions to run and when to continue or stop responding. 

\subsection{Data Retrieval Functions}

Table~\ref{tab:function_categories} summarizes the whitelisted functions available to RadOnc-GPT. When invoked, each function executes and returns outputs directly to the model. These outputs are designed to provide a coherent, structured representation of the underlying data, thereby maximizing the LLM’s ability to interpret and reason over the information. If no relevant data were found by a particular data retrieval function, each function explicitly indicates that no data could be found for the given inputs.

Notably, RadOnc-GPT does not employ conventional retrieval-augmented generation (RAG). In RAG, embedded documents are matched to queries via vector similarity and delivered back to the model as text chunks. This strategy was developed to surface relevant passages from vast, poorly organized, unstructured corpora such as the internet or a library, which often greatly exceed model context limits. By contrast, RadOnc-GPT leverages the fact that patient data within Epic is unstructured yet systematically organized and indexed. Clinical notes, for example, are timestamped, provider-signed, and tagged with metadata such as medical specialty, making them straightforward to retrieve without generic RAG pipelines. As LLMs continue to support increasingly larger contexts while inference costs per token decline (\$2.50 per 1 million input tokens for OpenAI's standard GPT-4o as of this report), the necessity of general-purpose RAG approaches further diminishes in medical settings such as RadOnc-GPT operates, where targeted, well-structured data retrieval suffices.

To avoid overwhelming the model with unnecessarily lengthy inputs, RadOnc-GPT employs a large set of highly specific retrieval functions. The guiding principle is that greater function granularity reduces the volume of data returned on a per-function basis. In rare cases where retrieved content exceeds the maximum allowable context length, a controlled pruning strategy is applied. The system prompt remains intact, while prior messages are selectively truncated. Because all retrieval functions return results in reverse chronological order (newest first, oldest last), pruning proceeds by removing fixed-size batches of words from the ends of the oldest messages. After each pass, the token count is recalculated; if the context remains too long, pruning continues until the message reaches a minimum threshold, at which point the process shifts to the next oldest message in the conversation. If all the messages have been pruned to the threshold and the conversation still exceeds the token budget, the threshold is lowered stepwise and pruning resumes. For this study, parameters were set as follows: maximum conversation history = 95,000 tokens, words removed per pass = 250, minimum threshold = 10,000 tokens, and threshold decrement = 2,500 tokens.

\begin{table}[h!]
\centering
\caption{RadOnc-GPT whitelisted function categories.}
\label{tab:function_categories}
\begin{tabular}{@{}p{3.2cm} p{4.8cm} p{6.5cm}@{}}
\toprule
\textbf{Category} & \textbf{Purpose / Data Retrieved} & \textbf{Representative Functions} \\ \midrule
Patient-oriented data retrieval &
Demographics, treatments, diagnoses, notes, reports, messages, appointments &
\texttt{get\_patient\_details}, \texttt{get\_patient\_treatment\_details}, \texttt{get\_patient\_diagnosis\_details}, \texttt{get\_patient\_clinical\_notes}, \texttt{get\_patient\_radiology\_reports}, \texttt{get\_patient\_pathology\_reports}, \texttt{get\_patient\_inbasket\_messages}, \texttt{get\_patient\_appointments}, \texttt{get\_physician\_appointments} \\[0.8em]

Clinical trials &
Identify recruiting trials and fetch eligibility criteria from clinicaltrials.gov&
\texttt{get\_list\_of\_clinical\_trials}, \texttt{get\_eligibility\_criteria} \\[0.8em]

Statistical information &
Generate cohort-level population statistics / search URLs &
\texttt{get\_patient\_population} \\[0.8em]

PSTAR (proton range) &
Compute proton stopping power, CSDA and projected ranges &
\texttt{get\_pstar\_data}, \texttt{get\_pstar\_material\_list} \\[0.8em]

PubMed &
Search biomedical literature, retrieve summaries and abstracts &
\texttt{pubmed\_search}, \texttt{pubmed\_summary}, \texttt{pubmed\_fetch} \\[0.8em]

DICOM &
Send DICOMs, monitor active streams, clear logs &
\texttt{send\_dicoms\_to\_server}, \texttt{get\_dicom\_streams\_info}, \texttt{clear\_dicom\_streams\_logs} \\[0.8em]

CTCAE (AE grading) &
List radiation-therapy adverse-event codes and grading details &
\texttt{get\_radiation\_therapy\_ae\_codes}, \texttt{get\_ctcae\_details\_by\_ae\_code} \\[0.8em]

Basic math &
Perform elementary arithmetic inside the agent loop &
\texttt{add}, \texttt{subtract}, \texttt{multiply}, \texttt{divide} \\ \bottomrule
\end{tabular}
\end{table}

\subsection{Two-Tier Evaluation Scaffold}
RadOnc-GPT's capability to extract patient outcomes was evaluated in two distinct tiers of escalating complexity. This two-tiered approach establishes foundational accuracy in basic retrieval tasks first, followed by more clinically relevant tasks of clinical outcomes labeling:

\begin{itemize}
    \item \textbf{Tier 1 (Data extraction from structured data)}: In the first evaluation, RadOnc-GPT was tasked with retrieving and reproducing structured data elements, including patient ID, name, sex, race, ethnicity, and treatment planning details such as course IDs, International Classification of Diseases (ICD) codes, plan IDs, and radiation types for each course. RadOnc-GPT’s outputs were directly compared with corresponding values extracted from the underlying database. Because this task relied exclusively on structured fields, no adjudication process was required, as the database entries were considered the ground truth.
    
    \item \textbf{Tier 2 (Patient outcomes labeling from structured and unstructured patient data)}: In the second evaluation, RadOnc-GPT was assessed on three clinically complex outcome-labeling tasks: (1) detection of mandibular osteoradionecrosis (ORN) in patients with HNC following radiation therapy, (2) detection of cancer recurrence in prostate cancer patients after radiation therapy, and (3) detection of cancer recurrence in HNC patients after surgery. These tasks required RadOnc-GPT to autonomously synthesize both structured data (e.g., demographics, treatment planning details, diagnosis codes) and unstructured data (e.g., clinical notes, radiology reports, pathology reports). For cancer recurrence detection, the same prompt design was applied across prostate cancer and HNC cohorts, allowing evaluation of RadOnc-GPT’s ability to generalize across disease sites and treatment modalities. The outcomes labels generated by RadOnc-GPT were compared with ground truth labels that were acquired through careful manual review by experienced physicians. Our study considers that ground truth may not be accurate. Each discrepancy between RadOnc-GPT and the ground truth was therefore adjudicated by an independent physician for a final result.
\end{itemize}


\subsection{LLM Task Orchestration}
\begin{figure}[ht]
  \centering
  \includegraphics[width=.80\textwidth]{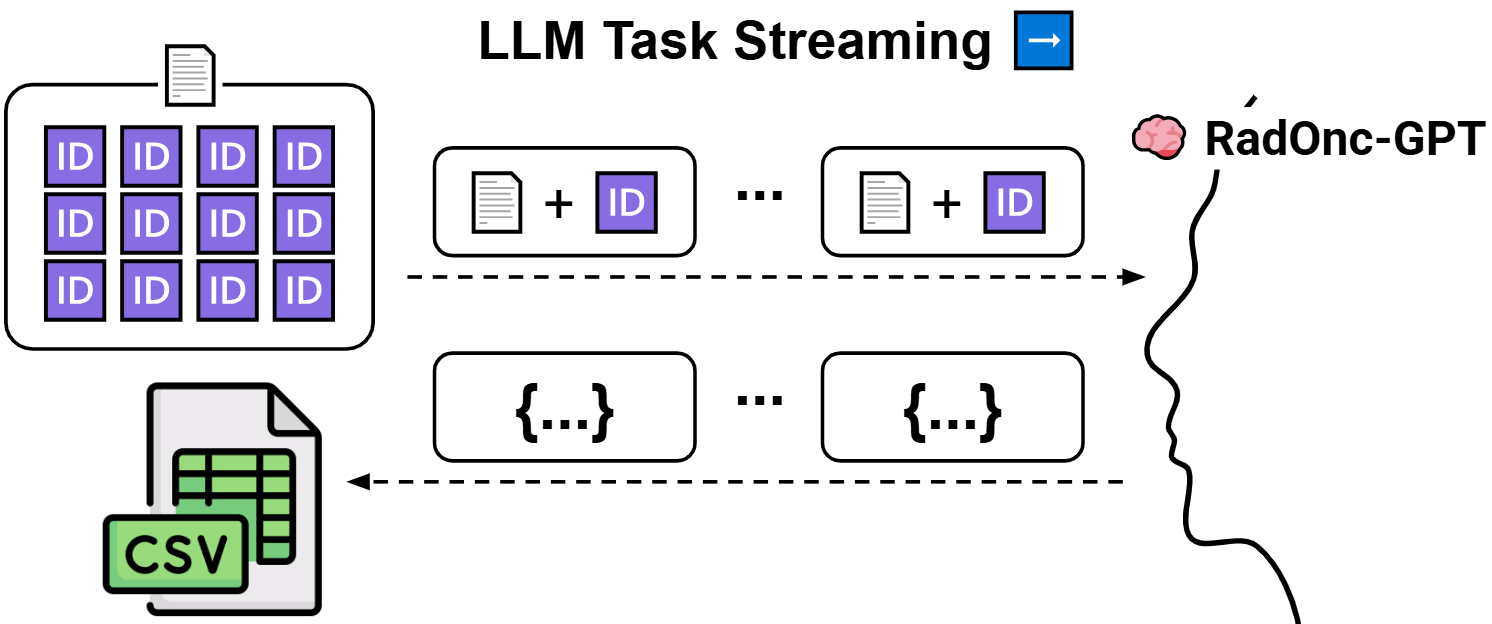}
  \caption{\textbf{LLM Task Streaming Orchestration.} External controller (LLM Task Streaming) sequentially supplying patient IDs and task instructions (prompts) to RadOnc-GPT, aggregating structured JSON objects.}
  \label{fig:task_streaming}
\end{figure}

To enable large-scale evaluation, we developed an external orchestration framework called LLM Task Streaming, designed to allow for processing a population of patients with RadOnc-GPT. See Figure~\ref{fig:task_streaming}. This system sequentially supplied RadOnc-GPT with patient IDs from each cohort along with task-specific instructions (prompts). The model’s outputs were returned as structured JSON objects, which were automatically parsed and consolidated into cohort-level CSV files for downstream analysis and physician adjudication. The framework supports parallel processing of patients, only limited by the artificial rate-limit associated with the account. In practice, data was processed in about 10–30 seconds per-patient depending on the volume of retrieved records.

\subsection{Prompts}

To execute a given task, RadOnc-GPT requires two inputs: the patient ID and a task-specific prompt. The full set of prompts used for each task are provided in the Appendix. In general, all prompts followed a consistent design strategy, specifying the overall task, data retrieval instructions, and task-specific instructions. These task-specific components could incorporate definitions, clinical criteria, or reasoning strategies (e.g., chain-of-thought guidance) to ensure accurate and consistent outputs. The process of developing the prompts involved frequent observation and feedback from experienced physicians.

\subsection{Cohorts and Ground Truth Labeling}

Four non-overlapping cohorts were selected for evaluation (\textbf{total $n=895$}). Table~\ref{tab:cohorts_summary} summarizes the number of patients (total, positive, and negative) included in the Tier 2 (complex) tasks. For the Tier 1 task, the cohort consisted of $n=500$ patients randomly selected from radiation therapy patients treated between 2015 and 2025. The ground truth for this evaluation was derived directly from the institutional Aria database.

\begin{table}[h!]
\centering
\caption{Patient cohort summary and baseline ground-truth label counts (pre-adjudication).}
\label{tab:cohorts_summary}
\begin{tabular}{@{}l r r r@{}}
\toprule
\textbf{Cohort} & \textbf{Total $n$} & \textbf{Positive (n)} & \textbf{Negative (n)} \\
\midrule
ORN detection (HNC) & 233 & 34 & 199 \\
Recurrence detection (Prost.) & 80  & 40 & 40  \\
Recurrence detection (HNC) & 82 & 48 & 34 \\
\bottomrule
\end{tabular}

\vspace{0.25em}
\begin{minipage}{0.92\linewidth}\footnotesize
Counts reflect baseline labels prior to adjudication; post-adjudication revisions are reported in the Results section.
\end{minipage}
\end{table}


\paragraph{Mandibular osteoradionecrosis (ORN) cohort.}
HNC patients were included according to the following criteria: adults ($>19$ years) treated between April 2013 and August 2019 with fraction sizes of 1.2–2.2 Gy and a total dose $>$30 Gy, who subsequently developed mandibular osteoradionecrosis (ORN) after their initial radiotherapy course. Patients with recurrent cancer were excluded. ORN status was established using CTCAE v4 osteonecrosis-of-the-jaw criteria and graded according to the Marx system. Positivity required documentation of ORN within the irradiated volume, supported by clinical notes, imaging, and lesion photographs. Alternative etiologies (e.g., infection, surgery, dental extraction, lymphoma, or disease outside the radiation field) were excluded.

\paragraph{HNC recurrence cohort.}
HNC patients (oropharynx, squamous cell carcinoma, prior radiation therapy) were selected from the Department of Otolaryngology Oropharynx RedCap registry. Cancer recurrence status was determined through chart review by an experienced clinical data abstractor ($>15$ years), supervised by a senior head-and-neck surgeon and a senior surgical oncologist. Ambiguities were resolved through weekly review sessions with the oncologist.

\paragraph{Prostate cancer recurrence cohort.}
Select prostate cancer patients were treated with definitive, curative-intent radiotherapy. Cancer recurrence was defined as any documented post-treatment progression of prostate cancer and was labeled if either criterion was met: (1) biochemical recurrence (PSA nadir + 2~ng/mL) or a clinician-determined, intervention-triggering PSA rise; and/or (2) clinical recurrence: local, regional, distant nodal, or metastatic progression confirmed by imaging or biopsy. Only prostate cancer recurrences were considered when other primary cancers were present.

\subsection{Adjudication Procedure}
For these complex outcomes labeling tasks (tier 2), discrepancies between RadOnc-GPT's results and baseline labels were independently adjudicated by oncologists with domain expertise specific to each clinical scenario (prostate cancer, HNC, mandibular osteoradionecrosis), one oncologist per dataset. Oncologists classified discrepancies explicitly into three categories:

\begin{enumerate}
    \item \textbf{RadOnc-GPT Correct}: Baseline label was determined erroneous (ground-truth error).
    \item \textbf{Baseline Label Correct}: RadOnc-GPT’s decision was erroneous (model error).
    \item \textbf{Indeterminate}: Insufficient evidence was available to reach a clear adjudication; these cases were excluded from final metrics.
\end{enumerate}

Accuracy, precision, recall, and F$_1$-score metrics were computed before and after the adjudication process.

\section{Results}

\subsection{Tier 1 Task - Structured Data Extraction (QA)}
In the initial structured extraction QA tier, RadOnc-GPT demonstrated exceptionally high fidelity in accurately retrieving structured demographic and treatment-related information. Across all 500 evaluated patients (totaling 3,000 individual demographic fields), RadOnc-GPT matched the demographic database records perfectly (100\%). Additionally, RadOnc-GPT accurately reproduced the treatment planning information in 497 out of 500 cases (99.4\%). These results established foundational trust in RadOnc-GPT's basic retrieval capability, a crucial prerequisite result before performing complex clinical outcomes labeling based on both structured and unstructured patient data.

\subsection{Tier 2 Tasks - Complex Clinical Outcomes Labeling based on Structured and Unstructured patient data}

\paragraph{Mandibular Osteoradionecrosis (ORN) in HNC Patients:}
Among 233 HNC patients evaluated for ORN, RadOnc-GPT initially achieved an accuracy of 84.5\%, with high recall (88.2\%) but lower precision (48.4\%). After adjudication by an independent radiation oncologist, overall accuracy significantly improved to 95.2\%. The adjudication process revealed that of the 36 initial discrepancies (32 false positives, 4 false negatives):

\begin{itemize}
    \item 25 cases (69\%) were previously unrecognized ground-truth errors, primarily due to outdated or incorrectly documented clinical information.
    \item 7 discrepancies were confirmed as genuine model errors.
    \item 4 cases were indeterminate due to insufficient evidence, and were excluded from final analysis.
\end{itemize}

Post-adjudication, RadOnc-GPT achieved 100\% recall, eliminating all false negatives.

\paragraph{Prostate Cancer Recurrence:}
In the prostate cancer recurrence cohort (80 patients), RadOnc-GPT initially reached an accuracy of 92.5\%, with precision and recall also both at 92.5\%. Post-adjudication, the accuracy improved to 95.0\%. Of the 6 initial discrepancies (3 false positives, 3 false negatives):

\begin{itemize}
    \item 2 cases (33\%) were found to be ground-truth errors, both involving small-volume lymph node recurrences that were initially missed during manual chart review.
    \item 4 discrepancies were confirmed as genuine model errors.
    \item No indeterminate cases were identified in this cohort.
\end{itemize}

The post-adjudication recall remained at 92.5\%, demonstrating reliable clinical utility for identifying prostate cancer recurrence.

\paragraph{HNC Recurrence:}
For the HNC recurrence cohort (82 patients), initial accuracy was 92.7\%, with a precision of 92.0\% and recall of 95.8\%. After independent adjudication, accuracy significantly increased to 96.3\%. The 6 initial discrepancies (4 false positives, 2 false negatives) were adjudicated as follows:

\begin{itemize}
    \item 3 cases (50\%) were determined to be ground-truth errors, reflecting incorrect or outdated documentation.
    \item 2 cases were genuine model errors.
    \item 1 case was indeterminate and thus excluded from the final analysis.
\end{itemize}

Final recall was excellent at 97.9\%, reflecting a clinically important reduction in false negatives.

\subsection{Summary of Adjudication Results Across Complex Patient Outcomes Labeling Tasks based on Structured and Unstructured Patient Data}

Across three complex patient outcomes labeling tasks (ORN in HNC, prostate cancer recurrence, HNC recurrence; total $n=395$ before adjudication), expert review of all discordant cases ($n=48$) found 30 (63\%) baseline labeling errors, 13 model errors, and 5 indeterminate cases. Post\textendash{}adjudication, false negatives fell from 9 to 4, and pooled (micro\textendash{}averaged) performance improved from Precision/Recall/F$_1$/Accuracy of 74.3/92.6/82.5/87.8\% to 90.3/97.0/93.6/95.4\% (Table~\ref{tab:adjudication_summary}). Note that post\textendash{}adjudication denominators exclude indeterminate cases for the affected patient cohorts.

\begin{table}[ht]
\centering
\caption{Confusion counts and performance metrics \emph{before} and \emph{after} adjudication. Totals are pooled (micro\textendash{}averaged) across tasks. Post\textendash{}adjudication $N$ excludes indeterminate cases (ORN: 4; HNC recurrence: 1).}
\label{tab:adjudication_summary}
\begin{tabular}{@{}l l r r r r r r r r r@{}}
\toprule
\textbf{Task} & \textbf{Adj.} & \textbf{$N$} & \textbf{TP} & \textbf{FP} & \textbf{FN} & \textbf{TN} & \textbf{Pr.(\%)} & \textbf{Re.(\%)} & \textbf{F$_1$(\%)} & \textbf{Ac.(\%)} \\
\midrule
\multirow{2}{*}{ORN HNC} 
  & Before & 233 & 30 & 32 & 4 & 167 & 48.4 & 88.2 & 62.5 & 84.5 \\
  & After  & 229 & 48 & 11 & 0 & 170 & 81.4 & 100.0 & 89.7 & 95.2 \\
\addlinespace[1pt]
\multirow{2}{*}{Rec. (Prost.)} 
  & Before & 80  & 37 & 3  & 3 & 37 & 92.5 & 92.5 & 92.5 & 92.5 \\
  & After  & 80  & 37 & 1  & 3 & 39 & 97.4 & 92.5 & 94.9 & 95.0 \\
\addlinespace[1pt]
\multirow{2}{*}{Rec. HNC} 
  & Before & 82  & 46 & 4  & 2 & 30 & 92.0 & 95.8 & 93.9 & 92.7 \\
  & After  & 81  & 46 & 2  & 1 & 32 & 95.8 & 97.9 & 96.8 & 96.3 \\
\midrule
\multirow{2}{*}{\textbf{Total}} 
  & Before & 395 & 113 & 39 & 9 & 234 & 74.3 & 92.6 & 82.5 & 87.8 \\
  & After  & 390 & 131 & 14 & 4 & 241 & 90.3 & 97.0 & 93.6 & 95.4 \\
\bottomrule
\end{tabular}
\end{table}

\section{Discussion}
This study investigated whether an autonomous, self-retrieving LLM agent can simultaneously curate patient outcomes and audit patient data quality within a realistic clinical setting. We adopted a two-tier evaluation strategy by design: first establishing fidelity in structured data retrieval before advancing to the more complex task of patient outcome labeling. In Tier 1, RadOnc-GPT successfully reproduced structured data alone, achieving exact matches for all demographic fields and accurately reproducing \emph{per-course} treatment information in 497 of 500 patients, including course identifiers, ICD codes, delivered plan identifiers, and radiation modality. Because these comparisons were made against structured database values from Aria (Varian Medical Systems, Palo Alto, CA), no manual review was required, making this approach well-suited for routine quality assurance. Importantly, this QA test also surfaced a systematic error early in our evaluation: course IDs with leading numbers were occasionally truncated by the LLM. Identifying this issue allowed us to implement corrections and strengthen the retrieval pathway. Thus, Tier 1 not only served as a validity checkpoint but also provided a mechanism for hardening the retrieval process before progressing to more complex evaluative tasks.

For Tier~2 tasks, after adjudication, recall reached 100\% for ORN detection, 92.5\% for prostate cancer recurrence detection, and 97.9\% for HNC recurrence detection. In the context of surveillance and registry curation, false negatives are typically unobserved since negatives are not typically reviewed, whereas false positives remain reviewable; thus, a recall-oriented operating point is preferable. The ORN finding---absence of false negatives---is particularly noteworthy, as missed late toxicities carry substantial clinical consequences. Two design choices likely contributed to this performance: (1) prompts explicitly requiring evidence both for and against each Marx stage, and (2) function-call--guided retrieval with task-specific temporal bounds, which maintains clinical context while filtering out irrelevant text. Together, these elements mitigate risks such as fragmentation of clinically linked facts across retrieval chunks and the limitations of generic embeddings, which often underperform in radiotherapy-specific domains.

Adjudication reframed model-versus-label disagreements. Of 48 initial discrepancies, 30 (63\%) were attributable to baseline labeling errors, 13 to model-generated errors, and 5 remained indeterminate. After excluding indeterminate cases, the adjusted accuracies of the original institutional labels were 95.2\% for ORN, 97.5\% for prostate cancer recurrence, and 96.3\% for HNC recurrence. Sutton et~al.\ reported a registry accuracy for HNC recurrence of 89.4\% with sensitivity 61\% and specificity 99\%, further noting that longer time-to-recurrence was associated with false negatives (median 1,007 vs.\ 414 days for true positives) \cite{SUTTON2024225}. Against this backdrop, RadOnc-GPT achieved higher recall while maintaining high accuracy, and it surfaced labeling errors that would otherwise have persisted. The findings suggest that autonomous agents can enhance completeness relative to conventional registry processes, while simultaneously generating an audit trail to support adjudication.

Throughput supports near-real-time use. With rate-limited parallelism and per-patient latencies of 10-30 seconds, most cohorts can be processed within a single day. Outputs are returned as structured JSON, enabling seamless integration into dashboards, analytics pipelines, and trial-matching platforms. This operational profile is important because registry accuracy is time-sensitive: delays in documentation contribute to missed events in conventional workflows \cite{SUTTON2024225}. A nightly or on-demand run can substantially shorten this gap. Notably, the same cancer recurrence detection prompt was applied to both prostate and HNC, suggesting that disease-site generalization is feasible with foundational LLM-based systems.

Limitations of this study include its reliance on single-center data and single-reviewer adjudication for each dataset; multi-reviewer panels could shift borderline determinations. Additionally, although inference costs of commercial LLMs have been and will likely continue reducing over time, institution-wide runs using an agentic approach remain cost-sensitive at present. These factors motivate future benchmarking against open-source models, exploration of caching strategies, and prompt optimization of cheaper models. Finally, although the agent generated rationale-rich outputs that facilitated adjudication, we did not evaluate human--AI conflict resolution policies in live operations; such mechanisms will be essential for routine deployment.

\section{Conclusions}

RadOnc-GPT demonstrates that a \emph{self-retrieving} GPT-4o agent can:

\begin{enumerate}
    \item flawlessly reproduce fundamental structured data, establishing trust in its retrieval chain;  
    \item label complex late-toxicity and cancer recurrence endpoints with near-expert performance;  
    \item simultaneously surface hidden errors in institutional databases, creating a virtuous cycle of model-assisted quality improvement.  
\end{enumerate}

Because the approach is prompt-driven and requires no model fine-tuning, it is readily extensible to other oncology centers and disease sites. We envision autonomous LLM agents functioning as continuous auditors, running nightly against new patients, updating outcomes registries in real time, and allowing clinicians to focus on patient care. The anticipated result is richer and timelier evidence to inform precision radiotherapy.


\section*{Author Contributions}
J.H. created and developed RadOnc-GPT and conducted the experiments, Y.H. helped with writing and editing and formulating the study, V.C, L.M., and K.V.A. helped with reviewing and curating patient data and reviewing the manuscript, F.M., D.R., A.F., S.S., D.M., S.K., B.S., Y.G., M.N.W., S.V., and N.L. helped review the manuscript, S.P., M.F., and M.B. helped with testing and reviewing the manuscript, R.F. helped with reviewing ORN data, M.B.O. helped with curating prostate patient data, D.E., M.W., and S.R.B. helped review/compare RadOnc-GPT outputs, and W.L. was the P.I and co-creator. 

\section*{Acknowledgement}
This study was reviewed and approved by the Mayo Clinic Institutional Review Board (IRB; protocol number 24-010276). The requirement for informed consent was waived by the IRB due to the retrospective nature of the study. All procedures were conducted in accordance with the ethical standards of the Mayo Clinic and with relevant national and international guidelines and regulations.

\section*{Conflict of Interest}
There are no conflicts of interest to report relating to this project.

\section*{Funding}
This research was supported by NIH/BIBIB R01EB293388, by NIH/NCI R01CA280134, by the Eric \& Wendy Schmidt Fund for AI Research \& Innovation, and by the Kemper Marley Foundation.

\bibliographystyle{vancouver}
\bibliography{references}

\appendix

\section{Prompts}

The following prompt was used for the structured QA task:

\begin{verbatim}
Retrieve the patient details and patient treatment details for patient 
{patient_id}. Based on the delivered treatment details, report the following
information, like this:

```
{
    "patient_id": "12345678",
    "first_name": "first name",
    "last_name": "last name (surname)",
    "sex": "male"/"female",
    "race": "put race here",
    "ethnicity": "put ethnicity here",
    "delivered_courses": [
        {
            "course_id": "course ID 1",
            "icd_codes": ["C61", "C75.1", ...],
            "delivered_plan_ids": ["plan ID 1", "plan ID 2", ...],
            "radiation_type": "proton" or "photon" or "electron"
        },
        {
            "course_id": "course ID 2",
            "icd_codes": ["C61", "C75.1", ...],
            "delivered_plan_ids": ["plan ID 1", "plan ID 2", ...],
            "radiation_type": "proton" or "photon" or "electron"
        },
    ]
}
```
\end{verbatim}

\noindent The following prompt was used for determining whether the patient experienced ORN:

\begin{verbatim}
Determine whether patient {patient_id} has ever experienced 
osteoradionecrosis (ORN). If the patient did experience ORN, 
then grade the severity (using the Marx staging system). 
Osteoradionecrosis (ORN) of the mandible refers to a condition 
occurring within the radiation treatment (RT) field in the head
and neck region, where mucosal breakdown or delayed healing 
leads to persistent exposure of the mandibular bone. Key criteria
include bone exposure lasting for at least three months, evidence
of necrotic (dead) bone, and the absence of recurrent tumor or 
metastases at the affected site. Additionally, ORN may present 
with radiographic evidence of bone necrosis even when the 
overlying mucosa remains intact. In order to determine whether 
the patient experienced ORN, first retrieve the patient details, 
patient diagnosis details, patient treatment details, radiology 
reports, pathology reports, and patient clinical notes (note_type 
of radiology, pathology, surgery, radiation_oncology, and ENT). 
Provide a detailed summary of the retrieved data, in the context 
of determining whether the patient experienced ORN, in a section 
called Patient History.

Next, in a section called Data Quantity, quantify the total 
amount of patient data that was retrieved by summing the 'number
of records count' from each of the retrieved patient data (not 
including patient details, patient diagnoses details, or patient
treatment details records).

Next, grade the severity of ORN in a section called Marx Staging.
The Marx staging system for grading ORN, proposed by Robert E. 
Marx in an article entitled "Osteoradionecrosis: A New Concept 
of Its Pathophysiology", is as follows: "Stage 1 is defined as 
exposed alveolar or mandibular bone without pathologic fracture.
If hyperbaric oxygen therapy (HBO) is used and the exposed bone
responds positively as a result, this is an indication of stage
1; Stage 2 is disease which does not respond to HBO therapy if 
it is given, and requires sequestrectomy and saucerization; and
Stage 3 is full-thickness bone damage or pathological fracture,
usually requiring complete resection and reconstruction with free
tissue." List any evidence for/against each stage (1, 2, or 3). 
If none of the criteria is met for any stage, as defined, then we 
shall define this as stage 0 (no evidence of ORN).

Finally, in a section called Concluding Remarks, make the case 
for or against the patient having ORN. After your concluding 
remarks, **provide the results using the following format**:

```
{
    'stage': '0, 1, 2, or 3', 
    'total number of records': '10, 14, 25, 115, etc.'
}
```
\end{verbatim}

\noindent The following prompt was used for determining whether the patient had cancer recurrence:

\begin{verbatim}
#### Task
I need your help in determining whether patient {patient_id} has 
had cancer recurrence at any point in their history relating to 
one of their prior cancer diagnoses.

#### Definition of Cancer Recurrence
According to the National Cancer Institute, recurrent cancer is 
cancer that has recurred (come back), usually after a period of 
time during which the cancer could not be detected. The cancer 
may come back to the same place as the original (primary) tumor 
or to another place in the body, but is only considered 
recurrence if it is related to the primary tumor. If it is a 
new tumor that is unrelated (a new, secondary primary tumor), 
then it is not recurrence relating to the original primary 
tumor. However, there is also the possibility that recurrence 
occurred relating to the secondary primary tumor.

#### Data Retrieval
To make a determination of whether the patient experienced cancer
recurrence in their history, you'll need to retrieve the following
data...

    * Patient treatment details
    * Radiation oncology notes (`note_type` = `radiation_oncology`)
    * Pathology notes (`note_type` = `pathology`)
    * Radiology notes (`note_type` = `radiology`)
    * Urology notes (`note_type` = `urology`... wait until you see
      diagnosis and only retrieve if the diagnosis is related to 
      urology)
    * ENT notes (`note_type` = `ent`... wait until you see diagnosis
      and only retrieve if the diagnosis is related to head and neck 
      cancers)
    * Radiology reports

When pulling any of the radiation oncology notes, the pathology 
notes, radiology notes, urology notes, ENT notes, and the radiology
reports, use the `date_minimum` input (I added `date_minimum` as a
new input for these data retrieval functions, format is %Y-%m-%d 
%H:%M:%S) and retrieve them all simultaneously using the last 
treatment date of the earliest treatment course as the 
`date_minimum`. Because you need the treatment details and the 
diagnoses details before retrieving certain patient data, you should
retrieve the treatment details and diagnoses details first. Go ahead
and retrieve them. Next, remark as to when the last treatment date 
of the first treatment course occurred, what the diagnoses are, and
whether the urology or ENT clinical notes are needed (based on 
whether the diagnoses are urology-related or head-and-neck-related
- note that prostate cancer is related to urology). After that, 
depending on the last treatment data and diagnosis, retrieve the 
rest of the patient data. If the patient does not have treatment 
details, then do not use the `date_minimum` input.
    
#### Instructions for Determining Cancer Recurrence
Before providing a yes/no answer on whether the patient had cancer
recurrence...

    * Summarize each dataset that was retrieved (including the 
      treatment details) in a section called "Summary of Retrieved 
      Data". Create a subsection for each summary.
    * Summarize the diagnoses that were treated noting the treatment 
      dates in a section called "Summary of Diagnoses".
    * Construct a timeline of important events that include the 
      diagnostic imaging reports, pathology imaging/lab reports, the
      actual diagnosis of the disease(s) with onset date(s), the 
      treatment of the disease(s), follow-up notes (note which
      diagnosis/treatment the follow-up is referring to (most likely
      the most recent one)) and any other important information 
      relating to the progression of disease over time or lack 
      thereof in a section called "Disease Timeline". You must 
      include PSA measurements in the timeline if the patient had 
      prostate cancer.
    * Next, we want to consider an argument that supports recurrence
      and an argument that does not support recurrence in a section 
      called "Argument For Recurrence/Argument Against Recurrence". 
      First, write a paragraph (starting with "The most plausible 
      argument supporting the case that cancer recurrence occurred in 
      their history is...") that describes the most plausible 
      explanation for how the cancer recurred in their history, 
      progressing from one diagnosis to another (consider spatial 
      proximity and time proximity). Next, write a paragraph (starting 
      with "The most plausible argument supporting the case that 
      cancer recurrence did not occur in their history is...") that 
      describes the most plausible explanation for how/why the cancer 
      did not recur in their history. You must write both of these 
      paragraphs unless there was only one diagnosis in their history
      (in which case recurrence was very unlikely).
    * Finally, explain your reasoning as to which choice is most 
      plausible, weighing the evidence, followed by a clear statement 
      of your final decision in a section called "Concluding Remarks".
    * Provide the answer in an "Answer" section using this JSON format:
```json
{{
    "recurrence": "yes/no"
}}
```
\end{verbatim}

\end{document}